\documentclass[letterpaper]{article} % DO NOT CHANGE THIS
\usepackage{aaai24}  % DO NOT CHANGE THIS
\usepackage{times}  % DO NOT CHANGE THIS
\usepackage{helvet}  % DO NOT CHANGE THIS
\usepackage{courier}  % DO NOT CHANGE THIS
\usepackage[hyphens]{url}  % DO NOT CHANGE THIS
\usepackage{graphicx} % DO NOT CHANGE THIS
\usepackage{natbib}  % DO NOT CHANGE THIS AND DO NOT ADD ANY OPTIONS TO IT
\usepackage{caption} % DO NOT CHANGE THIS AND DO NOT ADD ANY OPTIONS TO IT
\usepackage{algorithm}
\usepackage{algorithmic}

\usepackage{newfloat}
\usepackage{listings}
\usepackage{amssymb}
\usepackage{amsmath}
\usepackage{multirow}
\usepackage{colortbl} 
\usepackage{xcolor}
\usepackage{booktabs}
\usepackage{subcaption}
\usepackage{appendix}
\DeclareCaptionStyle{ruled}{labelfont=normalfont,labelsep=colon,strut=off} % DO NOT CHANGE THIS
\floatstyle{ruled}
\newfloat{listing}{tb}{lst}{}
\floatname{listing}{Listing}
\title{Set Prediction Guided by Semantic Concepts for Diverse Video Captioning}
\author{
Yifan Lu\textsuperscript{\rm 1,2}\equalcontrib,
Ziqi Zhang\textsuperscript{\rm 1}\equalcontrib,
Chunfeng Yuan\textsuperscript{\rm 1}\thanks{Corresponding author.},
Peng Li\textsuperscript{\rm 3,4},
Yan Wang\textsuperscript{\rm 3,4},
Bing Li\textsuperscript{\rm 1},
Weiming Hu\textsuperscript{\rm 1,2,5}
}
\affiliations{
\textsuperscript{\rm 1}State Key Laboratory of Multimodal Artificial Intelligence Systems, CASIA \\
\textsuperscript{\rm 2}School of Artificial Intelligence, University of Chinese Academy of Sciences \\
\textsuperscript{\rm 3}Alibaba Group 
\textsuperscript{\rm 4}Zhejiang Linkheer Science and Technology Co., Ltd.\\
\textsuperscript{\rm 5}School of Information Science and Technology, ShanghaiTech University \\
\{luyifan2021, zhangziqi2017\}@ia.ac.cn,
\{cfyuan, bli, wmhu\}@nlpr.ia.ac.cn,
\{sanjie.lp,  wy84378\}@alibaba-inc.com
}
\usepackage{bibentry}

\begin{document}

\maketitle

\begin{abstract}
	Diverse video captioning aims to generate a set of sentences to describe the given video in various aspects. Mainstream methods are trained with independent pairs of a video and a caption from its ground-truth set without exploiting the intra-set relationship, resulting in low diversity of generated captions. 
	Different from them, we formulate diverse captioning into a semantic-concept-guided set prediction (SCG-SP) problem by fitting the predicted caption set to the ground-truth set, where the set-level relationship is fully captured. Specifically, our set prediction consists of two synergistic tasks, i.e., caption generation and an auxiliary task of concept combination prediction providing extra semantic supervision. 
	Each caption in the set is attached to a concept combination indicating the primary semantic content of the caption and facilitating element alignment in set prediction.
	Furthermore, we apply a diversity regularization term on concepts to encourage the model to generate semantically diverse captions with various concept combinations. 
	These two tasks share multiple semantics-specific encodings as input, which are obtained by iterative interaction between visual features and conceptual queries. 
	The correspondence between the generated captions and specific concept combinations further guarantees the interpretability of our model. Extensive experiments on benchmark datasets show that the proposed SCG-SP achieves state-of-the-art (SOTA) performance under both relevance and diversity metrics.
\end{abstract}

\renewcommand{\arraystretch}{0.99}

\section{Introduction}
\begin{figure}[t]
	\begin{center}
		\includegraphics[width=0.90\linewidth]{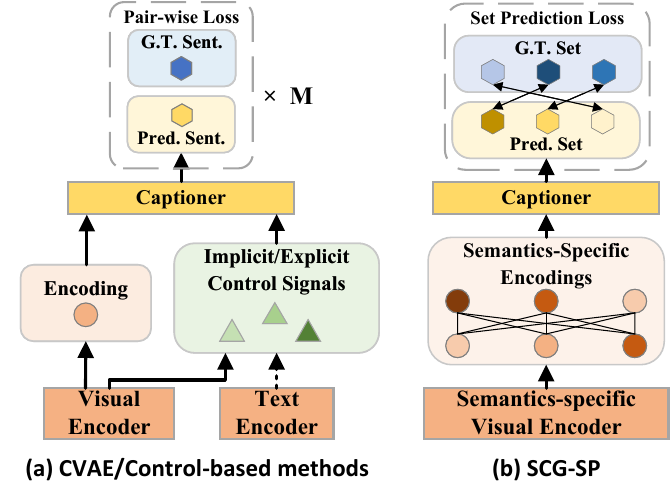}
	\end{center}
	\caption{Difference between (a) existing CVAE/control-based diverse captioning methods and (b) our proposed SCG-SP. 
		% CVAE-based and control-based methods generate sentences conditioned on a single visual encoding and multiple control signals.
		% Note that sampled latent variables can be regarded as implicit control signals in CVAE-based methods.
		There is no direct interaction among generated captions in CVAE-based or control-based methods, where the loss is calculated with independent training samples.
		% Therefore the sampling spaces of both implicit and explicit control signals do not model the intra-set relationships. The dotted line indicates not appearing at the inference stage. 
		Our proposed SCG-SP generates captions based on multiple semantics-specific visual encodings with sufficient interaction and is trained by a set-level prediction loss to exploit the set-level relationship.}
	\label{fig:1}
\end{figure}

% background

% div vc
Diverse video captioning (DivVC) is an emerging research branch of vision-language tasks. DivVC aims to generate a set of multiple captions that are semantically related to the given video and distinct from one another. DivVC overcomes the limitation of traditional video captioning methods, which only generate a single sentence for a video and fail to cover a wealth of visual information \cite{liu2022show, nie2022search}.

% Generating diverse sentences has been a hotspot in the field of image captioning (DivIC), while in the domain of video, limited number of researches focus on diverse captioning.

% CVAE
Mainstream diverse captioning methods can be classified into two categories: conditional variational encoder (CVAE) based methods \cite{aneja2019sequential, chen2019variational, deb2022variational, jain2017creativity, liu2022show, mahajan2020latent, mahajan2020diverse, nie2022search} and control-based methods \cite{chen2022learning, cornia2019show, deshpande2019fast}. As shown in Fig.\ref{fig:1}(a), both methods share a general pipeline that the diverse generation is conditioned on a visual encoding and multiple implicit or explicit control signals. 
% CVAE-based models apply a posterior latent distribution and a prior one, with the posterior taking both visual encoding and a encoded ground-truth sentence as inputs and the prior only taking visual encoding. By minimizing the KL-divergence between the posterior and the prior, the diversity learned by the posterior is transferred to the prior. At inference stage, 
% multiple latent variables are sampled conditioned on the prior and are fed to the caption decoder along with the a visual encoding to generate diverse captions. 
% CVAE-based methods apply a posterior latent distribution conditioned on both visual content and different sentences from the ground-truth set to encode diversity, which is transferred to the prior conditioned only on visual content when training. Multiple latent variables conditioned on the prior, regarded as implicit control signals, are sampled to generate diverse captions at the inference stage. Though with promising performance, CVAE-based methods are trained by pair-wise loss independently and ignore the relationship between the sentences for the same video. This results in the inadequate capture of diversity characteristics of captions. 

CVAE-based methods learn a latent distribution to encode diversity. Multiple latent variables, regarded as implicit control signals, are sampled from the learned distribution for diverse caption generation. 
However, CVAE-based methods are trained with a pair-wise loss, which ignores the intra-set relationship. This issue arises from the absence of direct interaction between generated captions for the same video and leads to inadequate capture of the diverse characteristics of captions.
Moreover, CVAE-based methods have limited interpretability because of the unrevealed relationship between the latent distribution and language patterns of the generated captions. 
Beyond CVAE-based methods, control-based methods apply explicit control signals to mitigate the problem of low interpretability. However, they still independently process the triplet of visual content, control signal, and caption, with the problem of ignoring intra-set relationship unsolved.

In this work, we propose a novel diverse video captioning model, \textbf{S}emantic-\textbf{C}oncept-\textbf{G}uided \textbf{S}et \textbf{P}rediction (SCG-SP).
To better exploit set-level relationships, we formulate diverse captioning as a set prediction problem. The target of DivVC is achieved by fitting the predicted caption set to the ground-truth caption set.  
To better grasp the intra-set diversity characteristics, we consider the source of caption diversity within a set. An important observation is that the diversity of captions depends on the differences in the semantic concepts they contain \cite{wang2019describing}.
These semantic concepts correspond to different objects, scenes, or actions in the visual content \cite{fang2015captions}. 
Since different combinations of concepts lead to different interpretations of rich visual content and are suitable for serving as guidance for caption generation, we incorporate concepts into set prediction for DivVC task.

% For better exploiting set-level relationships and achieving sentence generation with high relevance and diversity performance, we propose a novel model \textbf{S}emantic-\textbf{C}oncept-\textbf{G}uided \textbf{S}et \textbf{P}rediction (SCG-SP) for DivVC.
% The target of generating the sentence set with diverse and relevant captions can be achieved by fitting the predicted caption set to the ground-truth caption set. In this way, it is natural to formulate diverse captioning as a set prediction problem.
% how: 1 intra-set, interaction; 2 semantics 2.1 detected cpts 2.2 combination prediction
Specifically, as shown in Fig.\ref{fig:1}(b), the intra-set reasoning is achieved by interactions between multiple semantics-specific encodings and the set-level loss.
The encodings fully interact with each other, and each of them is decoded into a caption sentence accompanied by a combination of concepts. Therefore, our set prediction consists of two aspects, i.e., caption generation and the auxiliary concept combination prediction. For caption generation, we apply a GPT-2 \cite{radford2019language} captioner to leverage the vast knowledge learned from the external corpus.
The task of concept combination prediction is realized through multi-label classification, serving as an auxiliary task and providing semantic supervision.
% benifits: 1 set-level; 2 semantic, interpretable
SCG-SP is optimized by a set-level prediction loss composed of both captioning and classification costs. Through set-level reasoning, SCG-SP concerns the relationship within a set of diverse captions for the same video (intra-set) as well as the relationship among sets for various videos (inter-set), achieving considerable performance on both diversity and relevance. 
% Considerable performance on both diversity and relevance can be achieved through the sufficient exploitation of both intra-set and inter-set relationships. 
% 
Furthermore, each caption is guided by a particular concept combination, thus making the diverse generation interpretable. 

% We further incorporate semantic concept guidance that promotes set-level reasoning through video concept detection and concept combination prediction. SCG-SP encodes visual information in a semantics-specific way under the guidance of concepts and can generate sentences with high semantic diversity in a interpretable way. 

% method overview
% In practice, we apply a concept detector to predict the semantic concepts of the video. The GloVe embeddings \cite{pennington2014glove} of the top-\emph{M} concepts serve as the conceptual query inputs of the transformer-based concept-driven encoder. By performing self-attention within the queries and cross-attention with the frame-level features, we get multiple fully-interacted semantics-specific video encodings. Each encoding is decoded by a captioner and a classification head in the parallel decoding stage \cite{carion2020end}. 
% We apply a GPT-2 \cite{radford2019language} captioner to leverage the vast knowledge learned from the external corpus, making the captions more informative and diverse. Both captioning and classification costs contribute to the set prediction loss. 

% main contribution
The contributions of our work are listed as follows:
\begin{itemize}
	\setlength{\itemsep}{0pt}
	\item We propose a novel diverse video captioning model named SCG-SP to formulate DivVC as a set prediction problem. By exploiting intra-set and inter-set relationships, our model achieves high captioning performance in terms of both relevance and diversity.
	\item We further incorporate semantic concept guidance to promote set-level reasoning through concept detection and concept combination prediction, enabling SCG-SP to generate captions with high semantic diversity in an interpretable way. 
	\item Extensive experiments on MSVD, MSRVTT, and VATEX demonstrate that our proposed SCG-SP achieves state-of-the-art performances over existing methods.
\end{itemize}

\begin{figure*}
	\begin{center}
		%\fbox{\rule{0pt}{2in} \rule{.9\linewidth}{0pt}}
		\includegraphics[width=0.94\linewidth]{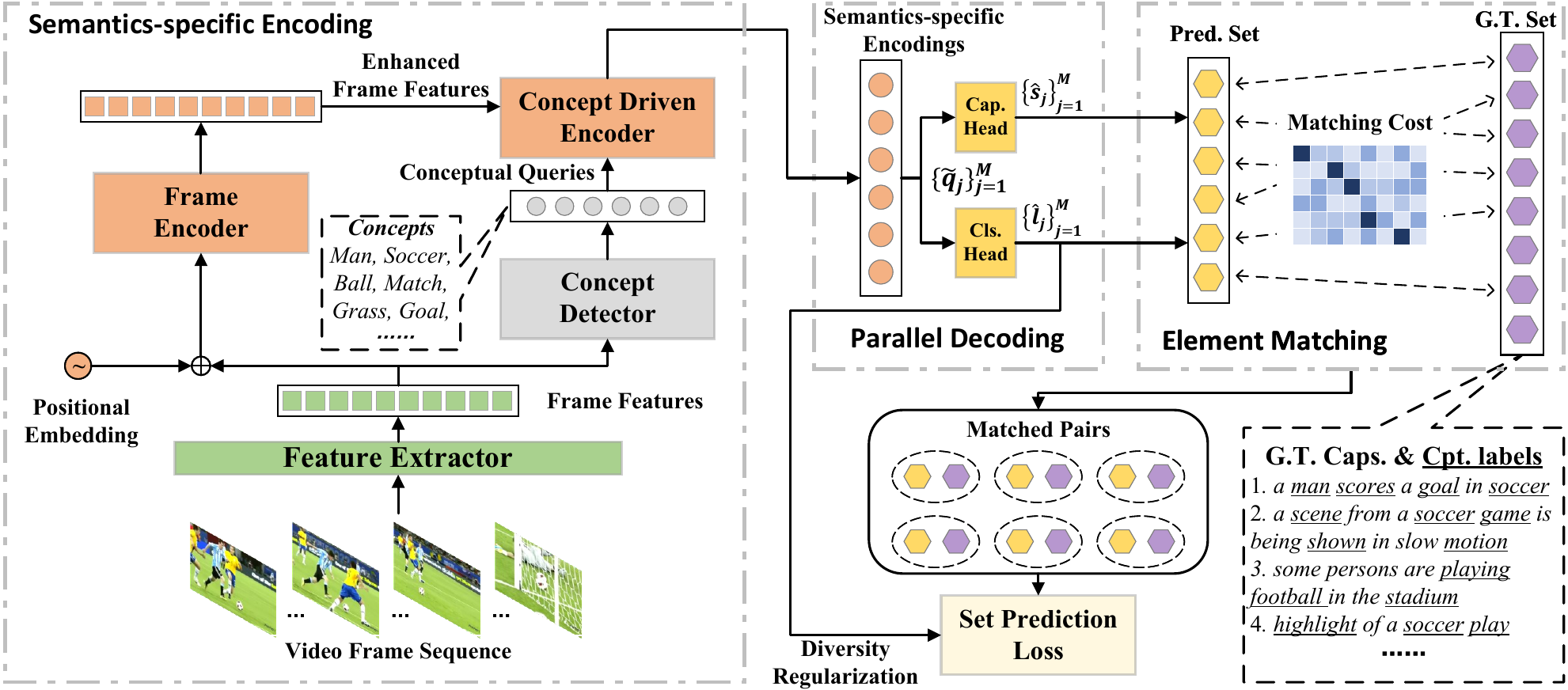}
	\end{center}
	\caption{  
		Overview of	the proposed SCG-SP. Based on pre-extracted video frame features, we first employ a temporal encoder, a concept detector, and a concept driven encoder to obtain multiple semantics-specific encodings for the input video. 
		In the parallel decoding stage, we apply a caption head and a classification head to respectively decode each encoding into a caption sentence and a concept combination label, which together form the prediction set. 
		By performing element matching between the predicted set and the ground-truth set, the set prediction loss is calculated over matched element pairs. Note that the ground-truth concept combination labels are assigned by taking nouns and verbs of high word frequency from the captions.}
	\label{fig:2}
\end{figure*}

\section{Related Works}
\subsubsection{Diverse Captioning}
% CVAE
Diverse image captioning (DivIC) has been an essential branch of image captioning. Some of the early methods are based on beam search \cite{vijayakumar2018diverse} and generative adversarial networks \cite{dai2017towards, li2018generating}. 
% CVAE-based methods apply a posterior latent distribution conditioned on both visual content and different ground-truth captions to encode diversity, which is transferred to the prior conditioned only on visual content. At the inference stage, multiple latent variables are sampled
% conditioned on the prior to generate diverse captions.
Recent DivIC methods \cite{aneja2019sequential, chen2019variational, jain2017creativity, mahajan2020latent, mahajan2020diverse} employ CVAE \cite{sohn2015learning} for diverse generations.
% CVAE-based models apply a posterior latent distribution and a prior one. The posterior is conditioned on both visual content and different ground-truth captions and can model the diversity, while the prior is only conditioned on visual encoding. By minimizing KL divergence between the posterior and the prior, the diversity learned by the posterior is transferred to the prior. At the inference stage, multiple latent variables are sampled conditioned on the prior and are fed to the decoder along with the visual encoding to generate diverse captions. 
% Beyond CVAE-based models, some works introduce control signals to make diverse generation controllable and interpretable, including POS \cite{deshpande2019fast} controlled by Part-of-Speech, SCT \cite{cornia2019show} controlled by image regions, and DML \cite{chen2022learning} controlled by learned discrete modes. 
Beyond CVAE-based models, Some studies \cite{deshpande2019fast, cornia2019show, chen2022learning} introduce control signals with explicit meaning, enabling diverse generations with interpretability.

In the domain of video, FLIP \cite{nie2022search} applies CVAE with latent prior modeled by a normalizing flow. VSLAN \cite{deb2022variational} employs a CVAE to generate diverse Part-of-Speech sequences for caption generation. 
SMCG \cite{yuan2020controllable} generates diverse captions controlled by exemplar sentences. 
However, these methods ignore the intra-set relationship, for there is no set-level loss or interaction among generated captions in these methods.
The recent-proposed STR \cite{liu2022show} takes the intra-set relationship into consideration for DivVC. STR clusters captions by topics and learns a latent distribution through two-stage training, respectively focusing on topics and paraphrasing. 
Note that our proposed SCG-SP performs set-level reasoning only requiring one-stage training, and is free of hand-crafted components like clustering, making the pipeline simple yet effective.

\subsubsection{Set Prediction}
Set prediction first emerges as a new paradigm for object detection. DETR \cite{carion2020end} applies a transformer encoder-decoder framework \cite{vaswani2017attention} to directly predict the set of objects in parallel. DETR is optimized by a set prediction loss based on the Hungarian algorithm \cite{kuhn1955hungarian} finding a unique element matching between the predicted and the ground-truth sets.
% , enforcing the element-permutation-invariance of the loss.
% By applying set prediction, DETR achieves promising performance and simplifies the detection pipelines by removing the need for hand-designed components. 
The set prediction framework has been extended to various multi-modal tasks, such as text-conditioned object detection \cite{kamath2021mdetr}, and spatio-temporal video grounding \cite{yang2022tubedetr}. 
PDVC \cite{wang2021end} uses DETR for describing videos with multiple events.
HMN \cite{ye2022hierarchical} applies a DETR-based module to predict a set of entities in video for single sentence caption. 
Our proposed SCG-SP applies the set prediction framework, which is consistent with the target of DivVC. Unlike HMN or PDVC, SCG-SP predicts a set of captions for a video with a single event.  
% Motivated by DETR, our proposed SCG-SP follows the set prediction framework, which is consistent with the target of DivVC, i.e. to generate a diverse caption set for a video. 

\subsubsection{Semantic Guidance}
Visual concepts corresponding to objects, scenes, or actions in the visual content can encode rich and high-level semantic cues \cite{fang2015captions}. Incorporating semantic concepts plays a vital role in vision-to-language tasks.
Previous works \cite{gan2017semantic, fang2022injecting, pan2017video, perez2021improving} detect semantic concepts from visual content leveraging Multiple Instance Learning (MIL) \cite{maron1997framework}. The predicted probability distribution of concepts is used as a high-level visual feature for further caption generation. 
% For example, LSTM-TSA \cite{pan2017video} and SemSynAN \cite{perez2021improving} apply a gated network to control the impact of concepts at generation; ViTCAP \cite{fang2022injecting} represents the semantic concepts at token level and fuses the concept token embedding with grid features for generation. 
% In our work, we apply semantic concept guidance by taking the embeddings of the detected concepts as conceptual queries for concept-driven video encoding. We further perform concept combination prediction at the decoding stage as an auxiliary task and provide extra semantic supervision.
Different from previous works, SCG-SP applies semantic concept to guide the Set Prediction process via two novel ways of conceptual queries and supervision from the classification task.

\section{Methodology}
% over view
% \subsection{Method Overview}
\subsection{Problem Formulation}
In the proposed SCG-SP, we formulate diverse video captioning as a set prediction problem. Given a video frame sequence $\mathcal{V}$, SCG-SP directly predicts a caption set with capacity $M$, which is denoted as $\hat{\mathcal{C}}=\{ \hat{c}_j | \hat{c}_j= (\hat{\mathbf{s}}_j, \hat{\mathbf{l}}_j) \}_{j=1} ^M$. 
For concept guidance, each caption sentence $\hat{\mathbf{s}}_j$ is attached with a concept combination label $\hat{\mathbf{l}}_j$.
% For concept guidance, each element in the predicted set has two attributes, not only a caption  $\hat{s}_j$ but a concept combination label $\hat{l}_j$ as well. In this way, the predicted set with capacity $M$ is denoted as $\hat{\mathcal{C}}=\{ \hat{c}_j | \hat{c}_j= (\hat{\mathbf{s}}_j, \hat{\mathbf{l}}_j) \}_{j=1} ^M$.
Similarly, the ground-truth caption set with capacity $M'$ is denoted as $\mathcal{C}=\{ c_j | c_j= (\mathbf{s}_j, \mathbf{l}_j) \}_{j=1} ^{M'}$.
The learning target is to fit the predicted set $\hat{\mathcal{C}}$ to the ground-truth set $\mathcal{C}$. Fig.\ref{fig:2} illustrates the architecture of SCG-SP, which consists of three main blocks: semantics-specific encoding, parallel decoding, and loss calculation. Details are discussed in the following subsections.   
% SCG-SP is trained by a set prediction loss. 

% Specifically, to generate the predicted set, we encode the video sequence guided by semantic concepts and apply parallel decoding for each semantics-specific encoding. 
% By applying element matching between the predicted set and the ground-truth set, matched pairs can be used to calculate the set prediction loss at the training stage. At the inference stage, the predicted diverse captions can be obtained through a single pass. Fig.\ref{fig:2} shows the architecture of SCG-SP. Details are discussed in the following subsections.

\subsection{Semantics-specific Encoding}
In the semantics-specific encoding stage, we represent an input video with multiple encodings $\tilde{\mathcal{Q}}$, each of which is subsequently decoded into a distinct caption. The encodings are derived from iterative attention on conceptual queries $\mathcal{Q}$ and temporal enhanced frame features $\tilde{\mathcal{F}}$, which infuse specific semantics into the encodings. The underlying interactions between encodings also facilitates the interactions between their corresponding generated captions.

\subsubsection{Temporal Encoding}
Given a sequence of video frames $\mathcal{V}$, we first employ pretrained 2D CNNs and 3D CNNs to extract the appearance and motion features of the video, respectively. By sampling keyframes and temporal concatenation, we get the frame features denoted as $\mathcal{F} = \{ \mathbf{f_i} \}_{i=1}^N$ for further encoding.
% We apply the encoder's architecture in transformer \cite{vaswani2017attention} for the Temporal Encoder. 
Then, we apply a Temporal Encoder composed of multiple self-attention layers. 
By performing self-attention on the frame feature sequence $\mathcal{F}$, the Temporal Encoder exploits inter-frame relationships and extract temporal-enhanced frame features denoted as $\tilde{\mathcal{F}}=\{ \tilde{\mathbf{f}}_i \}_{i=1}^N$.

\subsubsection{Concept Detection}
We design a Concept Detector to detect visual concepts in a video and generate conceptual queries $\mathcal{Q} = \{ \mathbf{q}_j \}_{j=1}^M$. We firstly build the concept vocabulary by selecting $N_c$ nouns or verbs with the highest word frequency. Subsequently, we tag each caption $\mathbf{s}_j$ with a 0/1 concept combination label $\mathbf{l}_j \in \mathbb{R}^{N_c}$, where $1$ is assigned to existing concepts and $0$ for non-existing ones. Finally, the pseudo ground-truth concept label of a video $\mathbf{l}_V$ is obtained by taking the bit-wise \emph{OR} operation: $\mathbf{l}_V = \mathbf{l}_1 \mid \mathbf{l}_2 \mid ... \mid \mathbf{l}_{M'}$, which means the $k$-th concept exists in the video if it appears in any of the $M'$ ground-truth captions. 

The Concept Detector performs multi-label classification task with $\mathbf{l}_V$ as the ground truth. It comprises a multi-layer perceptron (MLP) and predicts $\hat{\mathbf{l}}^V$, the probabilities of the $N_c$ concepts appearing in a video based on mean-pooled frame features.
The $M$ conceptual queries $\mathcal{Q} = \{ \mathbf{q}_j \}_{j=1}^M$ are obtained by taking the linear-projected GloVe embeddings \cite{pennington2014glove} of the $M$ concepts with the highest probabilities. Since the operation of taking the maximum is not differentiable, the detector is trained offline from the entire SCG-SP pipeline.

% The Concept Detector comprises a multi-layer perceptron (MLP), taking the mean-pooled frame features as input. It predicts a vector $\hat{\mathbf{l}}^V \in \mathbb{R}^{N_c}$, i.e., the probabilities of the $N_c$ concepts appearing in the video. The $M$ conceptual queries $\mathcal{Q} = \{ \mathbf{q}_j \}_{j=1}^M$ are obtained by taking the linear-projected GloVe embeddings \cite{pennington2014glove} of the $M$ concepts with the highest probabilities. 

% The Concept Detector is trained by optimizing a multi-label classification loss between $\hat{\mathbf{l}}_V$ and the ground truth $\mathbf{l}_V$. Since the process of taking the maximum is not differentiable, the concept detector is trained offline from the entire SCG-SP pipeline. 
% We build the concept vocabulary and annotate $\mathbf{l}_V$ following \cite{gan2017semantic}. $N_c$ nouns or verbs with the highest word frequency are selected as concepts. We tag each caption $\mathbf{s}_j$ with a one-hot concept combination label $\mathbf{l}_j \in \mathbb{R}^{N_c}$ where $1$ is assigned to existing concepts and $0$ for non-existing ones. For the video, $\mathbf{l}_V$ is obtained by taking the bit-wise \emph{OR} operation: $\mathbf{l}_V = \mathbf{l}_1 \mid \mathbf{l}_2 \mid ... \mid \mathbf{l}_{M'}$,
% % \begin{equation}
	% % 	\mathbf{l}_V = \mathbf{l}_1 \lor \mathbf{l}_2 \lor ... \lor \mathbf{l}_{M'},
	% % \end{equation}
% which means the ground-truth label for the $k$-th concept is set to $1$ if it appears in any of the $M'$ ground-truth captions.

\begin{figure}
	\begin{center}
		\includegraphics[width=\linewidth]{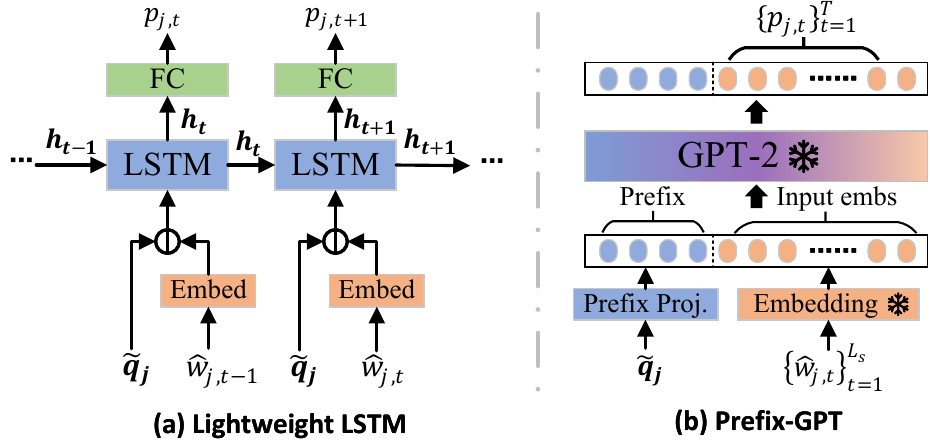}
	\end{center}
	\caption{Illustration of (a) the lightweight LSTM captioner and (b) the prefix-GPT captioner.
	}
	\label{fig:3}
\end{figure}

\subsubsection{Concept Driven Encoding}
% The Concept Driven Encoder takes the conceptual queries $\mathcal{Q}$ and the temporal-enhanced frame sequence $\mathcal{\tilde{F}}$ as input to generate semantics-specific encodings $\tilde{\mathcal{Q}}=\{ \tilde{\mathbf{q}}_j \}_{j=1}^M$. 
% The Concept Driven Encoder adopts transformer-decoder \cite{vaswani2017attention} architecture.
% First, the concepts are grouped into multiple combinations by performing self-attention within conceptual queries. Thus, each representation encodes a specific combination of concepts. 
% Then, in the cross-attention layer, concept-level representations serve as the query, while frame-level features of the video are employed as both the key and value to generate context-aware encodings that capture the relationship between specific concepts and video frames. Finally, the encodings are fully interacted through multiple self-attention layers.

The Concept Driven Encoder takes the conceptual queries $\mathcal{Q}$ and the temporal-enhanced frame sequence $\mathcal{\tilde{F}}$ as input to generate semantics-specific encodings $\tilde{\mathcal{Q}}=\{ \tilde{\mathbf{q}}_j \}_{j=1}^M$. 
Built on the transformer-decoder architecture \cite{vaswani2017attention}, it first groups concepts into combinations using self-attention on conceptual queries, resulting in representations encoding specific semantics. In the subsequent cross-attention layer, these concept-level representations act as queries, with frame-level video features serving as keys and values, to generate context-aware encodings that capture the relationship between specific concepts and video frames. The encodings are further refined through successive self-attention layers.

% Since the output query representations $\tilde{\mathcal{Q}}=\{ \tilde{\mathbf{q}}_j \}_{j=1}^M$ encode visual content in a semantics-specific way, $\tilde{\mathcal{Q}}$ is named as \emph{semantics-specific encodings}. Each encoding is decoded into an element in the predicted set in the parallel decoding stage. 

\subsection{Parallel Decoding}
% no need for counter
In the parallel decoding stage, the semantics-specific encodings are decoded into elements of the prediction set. A captioning head generates captions $\{\hat{\mathbf{s}}_j\}_{j=1} ^M$ for each encoding. Besides, a classification head predicts concept combination labels
$\{\hat{\mathbf{l}}_j\}_{j=1} ^M$, performing an auxiliary task that provides extra semantic supervision.
The two heads share the semantics-specific encodings as input.
%based on semantics-specific encodings
% for each element in the predicted set $\hat{\mathcal{C}}$.

\subsubsection{Captioning Head}
We apply two kinds of captioning heads to show the compatibility of our framework with both lightweight LSTM and large-scale pretrained GPT-2, as shown in Fig.\ref{fig:3}. 
Specifically, the $j$-th predicted caption $\hat{\mathbf{s}}_j$ consists of predicted words, i.e. $\hat{\mathbf{s}}_j = (\hat w_{j,1}, \hat w_{j,2}, \dots \hat w_{j,T})$. At time step $t$, $\hat w_{j,t}$ is conditioned on previous words $\hat w_{j,1:t-1}$ and the $j$-th encoding $\tilde{\mathbf{q}}_j$, i.e. $p(\hat w_{j,t}|\tilde{\mathbf{q}}_j, \hat w_{j,1:t-1})$ (denoted as $p_{j,t}$ for simplicity). $p_{j,t}$ is modelled by the captioning head as shown in eq.\ref{eq:1} and eq.\ref{eq:2} respectively for LSTM and GPT-2:
\begin{equation} 
	p_{j,t} = \text{Softmax} \Big( 
	\text{LSTM} \big(\big[\tilde{\mathbf{q}}_j,\text{Emb}(\hat{w}_{j,t-1})\big];\mathbf{h}_{j,t-1}\big) \Big),
	\label{eq:1}
\end{equation}
\begin{equation}
	p_{j,t} = \text{Softmax} \Big( 
	\text{GPT2} \big( \big[\text{MLP}(\tilde{\mathbf{q}}_j),\text{Emb}(\hat{w}_{j,1:t-1})\big] \big) \Big),
	\label{eq:2}
\end{equation}
where $[\cdot,\cdot]$ denotes concatenation; Emb is the word embedding layer; $\mathbf{h}_{j,t-1}$ denotes the previous hidden state of LSTM. The MLP projects $\tilde{\mathbf{q}}_j$ into word embedding space to serve as a prefix prompt. It is efficient to adapt the GPT-2 captioner to the captioning task by learning a prefix while freezing the GPT-2. 

\subsubsection{Classification Head}
% The classification head is composed of an MLP predicting the probability distribution of concepts, i.e., the concept combinations. For $j$-th element, the predicted $\hat{\mathbf{l}}_j \in \mathbb{R}^{N_c}$ is obtained by feeding $\tilde{\mathbf{q}}_j$ to the classifier.

The classification head is composed of an MLP. For the $j$-th element, it predicts the probability distribution of concepts $\hat{\mathbf{l}}_j \in \mathbb{R}^{N_c}$, i.e., the concept combinations, based on the encoding $\tilde{\mathbf{q}}_j$.

\subsection{Semantic Concept Guided Set Prediction Loss}
The proposed SCG-SP is optimized by a set prediction loss where the predicted and ground-truth set are treated as a whole to achieve set-level reasoning. The set prediction loss starts with a semantic concept guided element matching. The deterministic and optimal matching allows the loss to more accurately measure the differences between sets, aiding in better optimization. After that, the set prediction loss is calculated by pair. Besides caption cost, we consider classification loss of concept combination labels for extra semantic supervision. We further design a diversity regularization term on concepts to encourage the model to generate semantically diverse captions with various concept combinations. 

Specifically, the element matching finds the optimal element assignment $\hat\sigma$ between $\hat{\mathcal{C}}$ and $\mathcal{C}$ with the lowest matching cost based on concept combination labels. Assuming $\hat{\mathcal{C}}$ has a capacity no larger than $\mathcal{C}$, i.e., $M \leq M'$, each predicted element is assigned to a unique ground-truth element, i.e., $c_{\hat\sigma(j)}$ and $\hat{c}_{j}$ are matched into a pair. The optimal element assignment $\hat\sigma$ is obtained by using Hungarian algorithm \cite{kuhn1955hungarian} to solve:
\begin{equation}
	\hat\sigma = \arg\min_{\sigma \in \Omega_{M'}} \sum_{j=1}^{M} 
	L_{fl}	\big(\mathbf{l}_{\sigma(j)}, \mathbf{\hat{l}}_{j} \big),
	\label{eq:11}
\end{equation}
where $\Omega_{M'}$ is the searching space for $M'$ elements; $L_{fl}(\cdot, \cdot)$ refers to the focal loss \cite{lin2017focal}.

The captioning loss $L_{cap}$ is based on the Cross-Entropy loss of words:
\begin{equation}
	L_{cap} =
	-\sum_{j=1}^{M} \sum_{t=1}^{T} \delta(w_{\hat\sigma(j), t})^\top \log p_{j,t}
	,
\end{equation}
where $\delta(w_{\hat\sigma(j), t})$ denotes the one-hot vector for the $j$-th word in $\hat\sigma(j)$-th ground-truth caption; $L_s$ is the length of the sentence.
The classification loss is based on focal loss:
\begin{equation}
	L_{cls} = \sum_{j=1}^{M}
	L_{fl}	\big(\mathbf{l}_{\hat\sigma(j)}, \mathbf{\hat{l}}_{j} \big),
\end{equation}
The diversity regularization term $L_{div}$ is based on standard deviation (StdDev):
\begin{equation}
	L_{div} = -\sum_{k=1}^{N_c} 
	\text{StdDev} (\hat{l}_{1,k}, \hat{l}_{2,k}, \dots, \hat{l}_{M,k}), 
\end{equation}
where $\hat{l}_{j,k}$ is the probability of the $k$-th concept for the $j$-th element. Minimizing $L_{div}$ encourages SCG-SP to exploit more concept combinations. At last, the set prediction loss is obtained by taking the weighted sum of the captioning loss, the classification loss, and the diversity regularization term:
\begin{equation}
	L_{sp}= 
	L_{cap} + \lambda L_{cls} + \lambda_d L_{div}	
	.
	\label{eq:15}
\end{equation}

\section{Experiments}
In this section, we evaluate our proposed model on three public video datasets: MSVD \cite{chen2011collecting}, MSRVTT \cite{xu2016msr}, and VATEX \cite{wang2019vatex}, under both relevance and diversity metrics. We present the results compared with state-of-the-art models and report main ablation studies to demonstrate the effectiveness of our methods. More results are in \textbf{Appendix}.

\subsection{Experimental Setup}
\subsubsection{Datasets}
\textbf{MSVD} contains 1970 video from YouTube. Each video is annotated with 41 captions on average. We follow the split of 1200/100/670 for training, validation, and test. 
\textbf{MSRVTT} contains 10000 open domain videos. Each video is annotated with 20 captions. We follow the split of 6513/497/2990 for training, validation, and test.
\textbf{VATEX} contains 34991 videos, each with 10 English captions. We follow the split of 25991/3000/6000 for training, validation, and test.

\subsubsection{Preprocessing}
For each video, 8 frames/clips are sampled for feature extraction. We employ pre-extracted ResNet-101 \cite{he2016deep} and C3D \cite{hara2018can} features for MSVD and MSRVTT videos, and I3D \cite{carreira2017quo} for VATEX.
For captions sentences, we use NLTK toolkit \cite{bird2009natural} for part-of-speech tagging. We apply the 300-d GloVe embedding of detected concepts as conceptual queries. The size of the ground-truth caption set, $M'$, is set to 20/20/10 for MSVD/MSRVTT/VATEX. 

\subsubsection{Implementation Details} 
%The pre-extracted frame features are projected to 512 before being fed to the F-Encoder. 
The size of the concept vocabulary $N_c$ is set to 1000. The capacity of the predicted set and the conceptual query number $M$ is set to 20.
We implement two versions of our proposed model, i.e., SCG-SP-LSTM and SCG-SP-Prefix, using Lightweight LSTM and GPT-2 with prefix as captioners, respectively. For the Prefix-GPT captioner, we employ the smallest official version of GPT-2. The prefix length is set to 10. The weights of loss terms are set as $\lambda=1$ and $\lambda_d=0.5$. We apply AdamW as the optimizer. The learning rate and batch size are set to $8e^{-5}$ and 32 for SCG-SP-LSTM, $1e^{-5}$ and 8 for SCG-SP-Prefix. We use beam search with size 3 for generation at the inference stage. The model is implemented with PyTorch, and all the experiments are conducted on 1 RTX 3090 GPU.

\subsubsection{Evaluation Metrics}
The relevance of the captions is evaluated with metrics including BLEU@4 (B@4) \cite{papineni2002bleu}, METEOR (M) \cite{banerjee2005meteor}, ROUGE-L (R-L) \cite{lin2004rouge}, and CIDEr (C) \cite{vedantam2015cider}. We report \textbf{oracle} scores, where only the top-1 caption from each set is selected, demonstrating the upper-bound performance.
% In the practical test setting, the testing ground-truth captions are unavailable. Therefore, we also report \textbf{consensus} scores \cite{devlin2015exploring}. The predicted captions are reranked via the CIDEr score computed with the consensus references, which are retrieved with the testing video from the training corpus. We employ frozen-in-time \cite{bain2021frozen} as a video-to-text retriever. The consensus score takes the average of the top-5 reranked captions. 

The diversity of the captions is evaluated with metrics including Div-n (D-n), m-BLEU (m-B) \cite{aneja2019sequential}, and self-CIDEr (s-C) \cite{wang2019describing}. Div-n is the average ratio of distinct n-grams within each predicted caption set. We report Div-1 and Div-2 in our experiments. M-BLEU computes BLEU@4 for each diverse caption with the remaining captions in the set. Self-CIDEr calculates the ratio of the largest eigenvalue of the kernel matrix composed of CIDEr values between all pairs of captions in the set. \textbf{Higher} Div-n, self-CIDEr, and \textbf{lower} m-BLEU indicate more sentence diversity.

\begin{table}[t]
	\begin{center}
		\footnotesize
		\resizebox{\columnwidth}{!}{
			\setlength{\tabcolsep}{0.73mm}{
				\begin{tabular}{lcccccccc}
					\toprule[1.5pt]
					\multirow{2}{*}{Models} 
					% & \multirow{2}{*}{Ref.} & \multirow{2}{*}{Features} 
					& \multicolumn{4}{c}{MSVD} & \multicolumn{4}{c}{MSRVTT} \\
					\cline{2-9}
					% & & 
					& B@4 & M & R-L & C & B@4 & M & R-L & C \\
					\midrule[0.5pt]
					\multicolumn{9}{l}{\emph{\footnotesize top-1 over \textbf{20} sentences}} \\
					Div-BS$^\dag$ 
					% & AAAI 18 & RN-101, C3D 
					& 39.1 & 45.5 & 75.2 & 85.5 
					& 47.3 & 41.9 & 68.3 & 60.5\\
					SeqCVAE$^\ddag$
					% & ICCV 19 & InceptionRN-v2, I3D
					& 50.7 & 57.8 & 81.0 & 113.4 
					& 44.9 & 43.2 & 69.7 & 64.5 \\
					COSCVAE$^\ddag$
					% & NIPS 20 & InceptionRN-v2, I3D
					& 45.9 & 52.8 & 78.9 & 105.3 
					& 41.8 & 41.8 & 68.5 & 63.6 \\
					DML$^\dag$ 
					% & NIPS 22 & RN-101, C3D
					& 54.9 & 49.8 & 79.7 & 105.7 
					& 48.5 & 37.5 & 68.2 & 56.8 \\
					STR
					% & TMM 22 & InceptionRN-v2, I3D
					& 54.5 & 57.2 & 81.5 & 115.2 
					& 47.2 & 44.2 & 71.0 & 67.2 \\
					\rowcolor{lightgray} 
					SCG-SP-LSTM  
					% & Ours & RN-101, C3D
					& \textbf{58.4} & \textbf{60.5} & \textbf{82.9} & \textbf{120.2} 
					& \textbf{54.5} & 47.2 & \textbf{72.4} & 67.0 \\
					\rowcolor{lightgray} 
					SCG-SP-Prefix  
					% & Ours & RN-101, C3D
					& 56.5 & 58.9 & 81.0 & 114.7
					& 54.4 & \textbf{47.5} & 72.1 & \textbf{67.7} \\
					\midrule[0.5pt]
					\multicolumn{9}{l}{\emph{\footnotesize top-1 over \textbf{10} sentences}} \\
					VSLAN 
					% & WACV 22 & VGG16, RN-101, C3D, FR
					& \textbf{57.4} & 36.9 & 75.6 & 98.1 
					& 46.5 & 32.8 & 62.4 & 55.8 \\
					\rowcolor{lightgray}
					SCG-SP-LSTM  
					% & Ours & RN-101, C3D
					& 52.4 & \textbf{57.7} & \textbf{81.1} & \textbf{110.8}  
					& \textbf{48.1} & \textbf{43.8} & \textbf{69.8} & \textbf{61.7} \\
					\rowcolor{lightgray}
					SCG-SP-Prefix  
					& 51.6 & 56.1 & 79.7 & 109.9 
					& 46.4 & 43.4 & 69.2 & 61.5 \\
					\bottomrule[1.5pt]
				\end{tabular}
		}}
	\end{center}
	\caption{The oracle relevance scores on MSVD and MSRVTT. 
		% RN-101, RNX-101, InceptionRN-v2 and FR means ResNet-101, ResNeXt-101, Inception ResNet-v2 and Faster-RCNN. 
		\dag \ indicates DivIC models re-implemented for DivVC by us. \ddag \ indicates DivIC models re-implemented by authors of STR.}
	\label{tab:1}
\end{table}

\begin{table}
	% \begin{minipage}[t]{\columnwidth}
		\begin{center}
			\resizebox{\columnwidth}{!}{
				\setlength{\tabcolsep}{0.73mm}{
					\small
					\begin{tabular}{lcccccccc}
						\toprule[1.5pt]
						\multirow{2}{*}{Models} & 
						\multicolumn{4}{c}{MSVD} & 
						\multicolumn{4}{c}{MSRVTT} \\
						\cline{2-9}
						& D-1 
						& D-2
						& m-B$\downarrow$ 
						& s-C
						& D-1
						& D-2
						& m-B$\downarrow$ 
						& s-C \\
						\midrule[0.5pt]
						% ground-truth 
						% & 21.8 & 45.9 & 46.4 & 70.2 
						% & 37.5 & 69.0 & 25.3 & 84.0 \\
						% \midrule[0.5pt]
						\multicolumn{9}{l}{\emph{\footnotesize \textbf{20} sentences}} \\
						Div-BS 
						& 20.0 & 41.9 & \textbf{29.9} & \textbf{82.5} 
						&  22.1 & 42.0 & \textbf{49.4} & \textbf{80.9} \\
						DML 
						& 18.9 & 34.2 & 70.8 & 60.3
						& 20.1 & 36.1 & 69.9 & 63.0\\
						\rowcolor{lightgray}
						SCG-SP-LSTM
						& 21.0 & 33.7 & 63.9 & 54.1
						& 21.4 & 37.0 & 65.1 & 62.6\\
						\rowcolor{lightgray}
						SCG-SP-Prefix
						& \textbf{27.4} & \textbf{45.9} & \underline{50.2} & \underline{64.8}
						& \textbf{24.3} & \textbf{43.1} & \underline{58.4} & \underline{67.9}\\
						\midrule[0.5pt]
						\multicolumn{9}{l}{\emph{\footnotesize \textbf{10} sentences}} \\
						VSLAN 
						& 32.0 & 36.0 & 62.0 & -
						& 30.0 & 33.0 & 58.0 & -\\
						STR
						& 28.2 & 48.5 & 60.1 & -
						& 33.4 & \textbf{58.4} & \textbf{46.3} & - \\
						\rowcolor{lightgray}
						SCG-SP-LSTM
						& 25.7 & 39.0 & 62.6 & 54.5
						& 30.5 & 47.7 & 56.9 & 64.6 \\
						\rowcolor{lightgray}
						SCG-SP-Prefix
						& \textbf{33.3} & \textbf{52.4} & \textbf{58.4} & \textbf{67.9} 
						& \textbf{34.0} & 54.0 & 48.2 & \textbf{70.3} \\
						\bottomrule[1.5pt]
						
					\end{tabular}
			}}
		\end{center}
		%\vspace{-10mm}
		\caption{The diversity scores on MSVD and MSRVTT.}
		\label{tab:3}
	\end{table}
	% \vspace{3mm}
	% \end{minipage}

\subsection{Performance Comparison with SOTA}

To evaluate the effectiveness of SCG-SP, we compare our model with SOTA methods for DivVC, i.e., Div-BS \cite{vijayakumar2018diverse}, SeqCVAE \cite{aneja2019sequential}, COSCVAE \cite{mahajan2020diverse}, DML \cite{chen2022learning}, STR \cite{liu2022show}, and VSLAN \cite{deb2022variational}. Note that Div-BS, SeqCVAE, COSCVAE, and DML are re-implemented based on corresponding DivIC methods.
For a fair comparison with VSLAN and STR, we implement a version of SCG-SP with 10 generated captions per video on MSVD and MSRVTT. 

Tab.\ref{tab:1} shows the oracle relevance scores of DivVC models on MSVD and MSRVTT. On MSVD, SCG-SP-LSTM achieves the best performance on METEOR, ROUGE-L, and CIDEr. Though with lower BLEU@4, both SCG-SP-LSTM and SCG-SP-Prefix outperform VSLAN on the rest of the metrics by a large margin. On MSRVTT, our method achieves the best performance on all the metrics. 

Tab.\ref{tab:3} shows the diversity scores on MSVD and MSRVTT. 
Except for Div-BS, SCG-SP-Prefix scores the highest on MSVD and keeps the diversity on par with STR while outperforming the rest on MSRVTT. STR, as mentioned previously, also achieves set-level perception. The high diversity scores of STR and SCG-SP from the side prove the significance of the set-level relationship for diverse captioning. 
Note that Div-BS performs the best on two sentence-level diversity metrics because of its characteristic of generating all unique sentences. 
% However, due to the deficient relevance performance, the diversity scores of DivBS have little value of reference.
However, Div-BS performs poorly under relevance metrics.

Tab.\ref{tab:vatex} shows the relevance and diversity scores on VATEX. Our proposed SCG-SP-Prefix has the best overall performance.
To summarize, our model achieves considerable performance on the three benchmarks under both metrics, verifying the effectiveness of our proposed methods. 

\begin{table}
	% \begin{minipage}[t]{\columnwidth}
		\begin{center}
			\resizebox{\columnwidth}{!}{
				\setlength{\tabcolsep}{0.91mm}{
					\large
					\begin{tabular}{lcccccccc}
						\toprule[1.5pt]
						\multirow{2}{*}{Models} & \multicolumn{4}{c}{Relevance Scores} & \multicolumn{4}{c}{Diversity Scores} \\
						\cline{2-9}
						& B@4 & M & R-L & C  & D-1 & D-2 & m-B $\downarrow$ & s-C \\
						\midrule[0.5pt]
						Div-BS & 30.6 & 24.7 & 51.0 & 53.5 & 19.1 & \textbf{36.8} & 69.7 & \textbf{64.8} \\ 
						DML & \textbf{36.1} & 24.8 & 54.1 & 56.2 & 15.9 & 29.2 & 80.2 & 48.7 \\
						\rowcolor{lightgray}
						SCG-SP-LSTM & 32.0 & 25.8 & 52.8 & 56.4 & 10.4 & 18.5 & 91.5 & 52.4 \\ 
						\rowcolor{lightgray}
						SCG-SP-Prefix & \underline{34.5} & \textbf{26.1} & \textbf{54.1} & \textbf{61.2} & \textbf{24.4} & \underline{34.6} & \textbf{63.8} & \underline{60.1} \\
						\bottomrule[1.5pt]
					\end{tabular}
			}}
		\end{center}
		%\vspace{-10mm}
		\caption{Relevance and diversity scores on VATEX.}
		\label{tab:vatex}
		% \end{minipage}
\end{table}

\begin{table}
	\begin{center}
		\resizebox{0.75 \columnwidth}{!}{
			\setlength{\tabcolsep}{1.68mm}{
				\small
				\begin{tabular}{lcccccccc}
					\toprule[1.5pt]
					\multirow{2}{*}{Models} 
					% & \multirow{2}{*}{Ref.} & \multirow{2}{*}{Features} 
					& \multicolumn{4}{c}{MSRVTT} \\
					\cline{2-5}
					% & & 
					& B@4 & M & R-L & C \\
					\midrule[0.5pt]
					ORG-TRL
					& 43.6 & 28.8 & 62.1 & 50.9 \\
					SGN
					& 40.8 & 28.3 & 60.8 & 49.5 \\
					Open-Book
					& 42.8 & 29.3 & 61.7 & 52.9 \\
					SemSynAN
					& 46.4 & 30.4 & 64.7 & 51.9 \\
					% O2NA
					% & 41.6 & 28.5 & 62.4 & 51.1 \\
					HMN
					& 43.5 & 29.9 & 62.7 & 51.5 \\
					TextKG
					& 43.7 & 29.6 & 62.4 & 52.4 \\
					% UniVL
					% & 41.8 & 28.9 & 60.8 & 50.0 \\
					MV-GPT$^\S$
					& 48.9 & 38.7 & 64.0 & 60.0 \\
					SwinBERT$^\S$
					& 45.4 & 30.6 & 64.1 & 55.9 \\
					Vid2Seq$^\S$
					& - & 30.8 & - & 64.6 \\
					\rowcolor{lightgray} 
					SCG-SP-LSTM  
					& \textbf{54.5} & 47.2 & \textbf{72.4} & 67.0 \\
					\rowcolor{lightgray} 
					SCG-SP-Prefix  
					& 54.4 & \textbf{47.5} & 72.1 & \textbf{67.7} \\
					\bottomrule[1.5pt]
				\end{tabular}
		}}
	\end{center}
	\caption{Comparison with the SOTA methods for single sentence video captioning on MSRVTT. $\S$ indicates models pretrained on large-scale datasets}
	\label{tab:2}
\end{table}  

We also compare our model with SOTA methods for traditional single sentence video captioning, including ORG-TRL \cite{zhang2020object}, SGN \cite{ryu2021semantic}, OpenBook \cite{zhang2021open},  SemSynAN \cite{perez2021improving}, HMN \cite{ye2022hierarchical}, TextKG \cite{gu2023text}, MV-GPT \cite{seo2022end}, SwinBERT \cite{lin2022swinbert}, and Vid2Seq \cite{yang2023vid2seq}, as shown in Tab.\ref{tab:2}. 
For our model, we use oracle relevance scores derived from the best of 20 generated captions per video. For traditional models, scores are calculated using the single predicted caption. The significant improvement in scores of our model demonstrates its superior ability to generate high-quality captions, even when compared to models pretrained on large-scale data. This advantage is attributed to the fact that learning to describe a video with multiple sentences prevents mode collapse, as discussed in \cite{chen2022learning}.

\begin{table}
	\begin{center}
		\resizebox{\columnwidth}{!}{
			\setlength{\tabcolsep}{0.63mm}{
				\small
				\begin{tabular}{lcccccccc}
					\toprule[1.5pt]
					\multirow{2}{*}{Configurations} & \multicolumn{4}{c}{Relevance Scores} & \multicolumn{4}{c}{Diversity Scores} \\
					\cline{2-9}
					& B@4 & M & R-L & C  & D-1 & D-2 & m-B $\downarrow$ & s-C \\
					\midrule[0.5pt]
					LSTM-Base 
					& 51.1 & 45.4 & 71.2 & 64.7
					& 17.3 & 29.3 & 75.9 & 53.9 \\
					\footnotesize - w/o cpt. queries  
					& 49.7 & 45.3 & 70.7 & 63.5 
					& 13.8 & 22.8 & 86.5 & 46.2 \\
					\footnotesize - w/o cls. head 
					& 38.5 &  40.2 & 66.8 & 56.7
					& 8.4 & 11.1 & 96.2 & 19.8 \\
					\footnotesize - w/ $L_{div}$ 
					& \textbf{54.5} & \textbf{47.2} & \textbf{72.4} & \textbf{67.0}
					& \textbf{21.4} & \textbf{37.0} & \textbf{65.1} & \textbf{62.6} \\
					\midrule[0.5pt]
					Prefix-Base 
					& 51.6 & 46.3 & 71.2 & 65.9
					& 22.5 & 39.8 & 63.1 & 64.4 \\
					\footnotesize - w/o cpt. queries 
					& 49.8 & 46.5 & 70.9 & 66.2
					& 16.9 & 28.1 & 82.2 & 49.8\\
					\footnotesize - w/o cls. head  
					& 48.8 & 45.2 & 70.0 & 64.3 
					& 18.6 & 31.1 & 75.3 & 54.3 \\
					\footnotesize - w/ $L_{div}$ 
					& \textbf{54.4} & \textbf{47.5} & \textbf{72.1} & \textbf{67.7} 
					& \textbf{34.0} & \textbf{54.0} & \textbf{48.2} & \textbf{70.3} \\			
					\bottomrule[1.5pt]
				\end{tabular}
		}}
	\end{center}
	\caption{Ablative results on Semantic Concept Guidance.}
	% \vspace{-3mm}
	\label{tab:4}
\end{table}
\begin{figure}[t]
	\begin{center}
		\includegraphics[width=\linewidth]{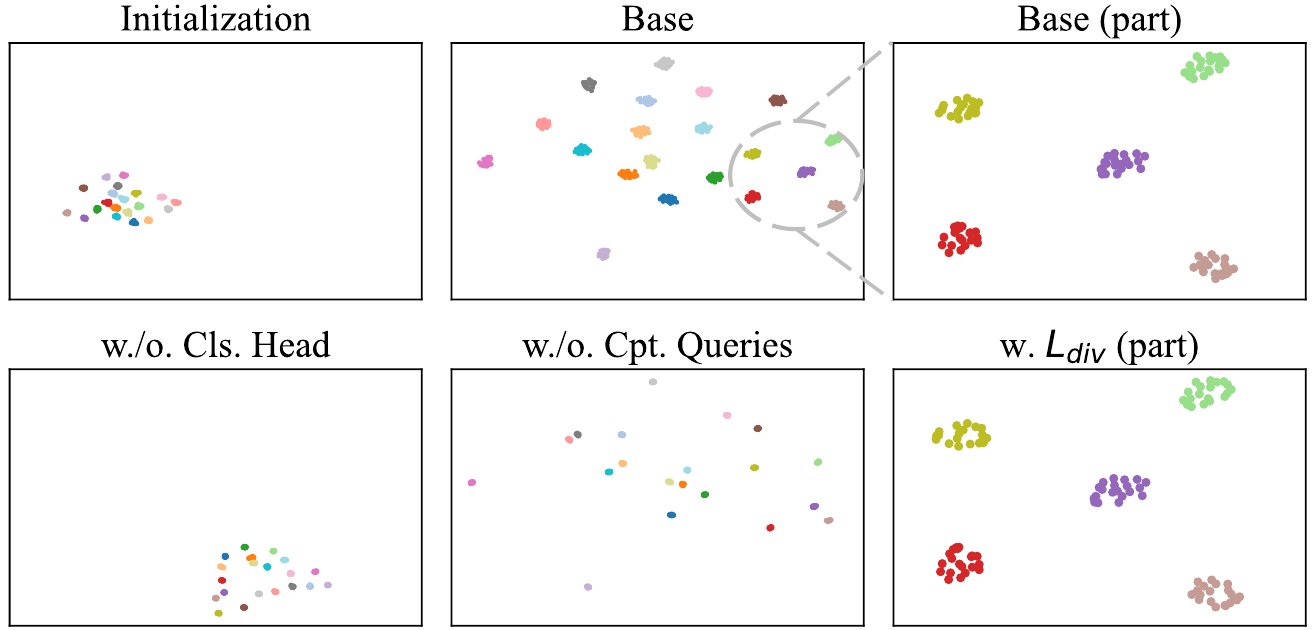}
	\end{center}
	\caption{Distribution of semantics-specific encodings. Different colors stand for encodings of different videos. We take the same part from the distribution maps of \emph{Base} and \emph{w/ $L_{div}$} for clear comparison. Best viewed in color.}
	\label{fig:4}
\end{figure}

\begin{figure*}[t]
	\begin{center}
		\includegraphics[width=\linewidth]{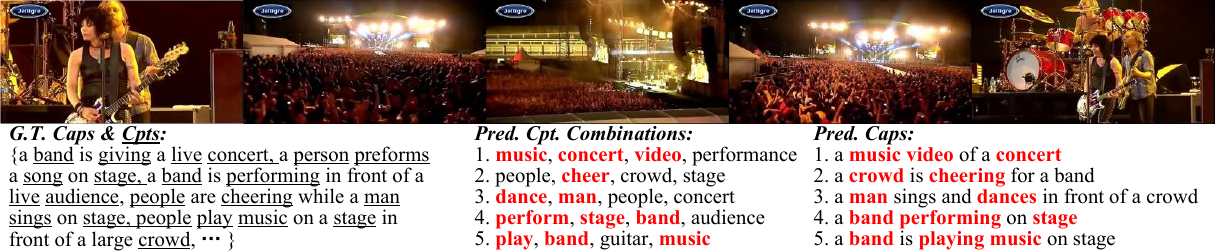}
	\end{center}
	\caption{An example of generations by SCG-SP-Prefix on MSRVTT. 
		% Predicted concepts with high probabilities are listed. 
		Concepts both showed in predicted combinations and captions are highlighted in red. Best viewed in color.
	}
	\label{fig:6}
\end{figure*}

\subsection{Ablation Studies}
\subsubsection{Evaluation on Semantic Concept Guidance} 
As shown in Tab.\ref{tab:4}, we design ablative experiments on MSRVTT to evaluate the three components of Semantic Concept Guidance (SCG).
The \emph{Base} configuration is obtained by removing the diversity regularization term $L_{div}$ from the full model. The other three settings are obtained by replacing conceptual queries with random vectors, removing the classification head, and adding diversity regularization, respectively.
Quantitative results show that for two versions, all three components of SCG bring performance gain on both relevance and diversity performance of diversity caption.

We further present the explanation through visualization of semantics-specific encodings distribution shown in Fig.\ref{fig:4}. We randomly sample 20 videos and generate 20 sets of encodings with models (LSTM version) trained under the four configurations. We plot t-SNE \cite{van2008visualizing} reduced encodings in different colors representing different sets. The initialized encodings are also plotted. The inter-set distance illustrates the distinction between videos with different visual content and represents the relevance performance of the model, while the intra-set distance indicates the diversity performance.
Therefore, the distribution of the encodings is highly related to the quantitative performance of relevance and diversity. 

As shown in Fig.\ref{fig:4}, initialized encodings and encodings by the model without classification head are with lower intra-set and inter-set distances compared with the encodings by the base model. Adding a diversity regularization term $L_{div}$ can increase the intra-set distance.
The model without conceptual queries can produce encodings with appropriate inter-set distance but low intra-set distance. 

% general recap
Combining both quantitative and qualitative results, we conclude that SCG improves diverse captioning performance by providing distinctive encoding sets and unique encodings within each set. Specifically, it is achieved by three means: 1) semantic information in conceptual queries; 2) deterministic element alignment and extra semantic supervision from concept combination prediction; 3) diversity preference brought by concept-based regularization.

% \begin{table}
	% 	\begin{center}
		% 		\resizebox{\columnwidth}{!}{
			% 			\setlength{\tabcolsep}{1.2mm}{
				% 				\small
				% 				\begin{tabular}{lccccccccc}
					% 					\toprule[1.5pt]
					% 					\multirow{2}{*}{Captioner} & Matching & \multicolumn{4}{c}{Relevance Scores} & \multicolumn{4}{c}{Diversity Scores} \\
					% 					\cline{3-10}
					% 					& Methods & B@4 & M & R-L & C  & D-1 & D-2 & m-B $\downarrow$ & s-C \\
					% 					\midrule[0.5pt]
					% 					\multirow{2}{*}{LSTM}
					%                         & Random & 36.7 & 39.3 & 65.9 & 55.0 & 7.9 & 10.3 & 96.5 & 17.3 \\ 
					%                         & Hungarian & 51.1 & 45.4 & 71.2 & 64.7 & 17.3 & 29.3 & 75.9 & 53.9 \\
					%                         \midrule[0.5pt]
					%                         \multirow{2}{*}{Prefix-GPT}
					%                         & Random & 39.9 & 40.0 & 66.7 & 58.6 & 10.1 & 14.4 & 91.0 & 28.9 \\ 
					%                         & Hungarian & 51.6 & 46.3 & 71.2 & 65.9	& 22.5 & 39.8 & 63.1 & 64.4 \\
					% 					\bottomrule[1.5pt]
					% 				\end{tabular}
				% 		}}
		% 	\end{center}
	% 	%\vspace{-10mm}
	% 	\caption{Ablative results on methods of element matching.}
	% 	\label{tab:hun}
	% \end{table}

\begin{table}
	\begin{center}
		\resizebox{0.95\columnwidth}{!}{
			\setlength{\tabcolsep}{1.85mm}{
				\small
				\begin{tabular}{lccccc}
					\toprule[1.5pt]
					Training style & C & m-B $\downarrow$ & s-C & $\#$ Params & SPE \\
					\midrule[0.5pt]
					From scratch & 64.3 & 72.9 & 57.1 & 169.1M & 424   \\
					Fine-tuning & \textbf{66.3} & 68.3 & 60.3 & 169.1M & 424 \\
					\midrule[0.5pt]
					Prefix-tuning & 65.9 & \textbf{63.1} & \textbf{64.4} & 44.6M & \textbf{325} \\
					\bottomrule[1.5pt]
				\end{tabular}
		}}
	\end{center}
	\caption{Ablative results on GPT-2 training style.}
	\label{tab:5}
\end{table}

\subsubsection{Analysis on Captioners} As shown in Tab.\ref{tab:1}-\ref{tab:vatex}, both SCG-SP-LSTM and SCG-SP-Prefix outperform SOTA methods. 
These results show the compatibility of our proposed SCG-SP framework for different captioners.

We also find that SCG-SP-Prefix outperforms the SCG-SP-LSTM under diversity metrics. For relevance metrics, though the Prefix version scores lower than the LSTM one on MSVD, the gap narrows on MSRVTT, and the Prefix version even scores better on VATEX.
These results demonstrate that as the complexity of the evaluation dataset increases, the advantages of using external knowledge become more pronounced.

By comparing the two halves in Tab.\ref{tab:4}, the ablative results of the two versions show consistency regarding effectiveness on SCG components.
However, the Prefix version has a lower requirement for conceptual guidance. For example, the LSTM version drops 8.0 and 34.1 on CIDEr and self-CIDEr, respectively, after removing the classification head, but for the Prefix version, the dropping is 1.6 and 10.1. We can also attribute this to external knowledge.

\subsubsection{Training of GPT} We employ three training styles for SCG-SP-Prefix, including training the GPT-2 model from scratch (i.e., without loading pretrained parameters), fine-tuning based on pretrained parameters, and prefix-tuning. For models trained by three styles within 100 epochs, captioning metrics on MSRVTT are listed with the number of trainable parameters and seconds per training epoch (SPE) in Tab.\ref{tab:5}. It is pretty difficult to fully train a GPT-2 from scratch with limited data and time. Thus the model trained from scratch performs the worst. Though the model with fine-tuned GPT-2 performs the best on CIDEr, the model with prefix-tuning, which is proven effective, achieves the highest diversity with  $73.6\%$ less trainable parameters $23.3\%$ less training time and only 0.4 drop on CIDEr. The results prove the efficiency of prefix-tuning.

\subsection{Qualitative Analysis}
% As discussed in section 4.4, we present a visualization of semantics-specific encodings in Fig.\ref{fig:5} to prove the effectiveness of semantic concept guidance.

We present an example of generations by our proposed SCG-SP-Prefix in Fig.\ref{fig:6}. The generated captions are with considerable diversity and are highly related to the predicted concept combinations. The results
demonstrate the effectiveness of guiding diverse generations with different combinations of concepts and 
show that the diverse generation of the proposed SCG-SP is highly interpretable. More examples are given in \textbf{Appendix}.

\section{Conclusion}
In this paper, we have proposed a novel model for diverse video captioning named SCG-SP, which stands for semantic-concept-guided set prediction. Through set-level reasoning, SCG-SP has captured the linguistic characteristics of the caption corpus. 
We have further incorporated semantic concept guidance of concept detection and concept combination prediction to improve the semantic diversity of captions and achieve interpretable generation.
Our proposed model has achieved the state-of-the-art performance on MSVD, MSRVTT, and VATEX benchmarks. Extensive quantitative and qualitative experiments have demonstrated the effectiveness of our methods.

\section{Acknowledgments}
This work is supported by National Key R\&D Program of China (No. 2022ZD0118501), Beijing Natural Science Foundation (Grant No. JQ21017, L223003, 4224093), the Natural Science Foundation of China (Grant No. 61972397, 62036011, 62192782, U2033210, 62202470).

\bibliography{aaai24}
\newpage
\mbox{}
\newpage

\appendix
\section{Appendix A: Additional Quantitative Results}

\subsection{Consensus Evaluation for Relevance Performance}
In the practical setting of relevance performance evaluation, the test ground-truth captions are not provided, making oracles scores unavailable. Therefore, we apply \textbf{consensus} evaluation \cite{devlin2015exploring}, which does not rely on the test ground truth. The predicted captions are reranked via the CIDEr score computed with the consensus references, which are captions from the training corpus and are related to the test video. We employ frozen-in-time \cite{bain2021frozen} as a video-to-text retriever to collect consensus references for each test video. The consensus score takes the average of the top-5 reranked captions. Tab.\ref{tab:consensus} shows the consensus-reranked relevance scores. SCG-SP-LSTM achieves the best performance in terms of four metrics on MSVD and three metrics on MSRVTT, further proving the superiority of our method.

\subsection{Performance with Stronger Feature Extractor and Captioner}
We use stronger vision encoder and captioner. As shown in Table \ref{tab:llama}, replacing ResNet101 \cite{he2016deep} backbone with \emph{CLIP-ViT-L} \cite{radford2019language} results in a performance gain of \textbf{+27.9} for CIDEr and \textbf{+9.4} for self-CIDEr. Further replacing GPT-2 decoder with \emph{Sheared-LLaMA-1.3B} \cite{xia2023sheared} brings a performance improvement of \textbf{+10.7} for CIDEr and \textbf{+4.0} for self-CIDEr, with our model attaining scores of \textbf{153.4} and \textbf{78.2}, on MSVD. 

We also compare with BLIP \cite{li2022blip} and BLIP-2 \cite{li2023blip}. We apply BLIP/BLIP-2 on DivVC by taking the temporally-attended vision encoding of sampled video frames as the text decoder’s conditional input and generating diverse captions through group beam search \cite{vijayakumar2018diverse}. As shown in Table \ref{tab:llama}, our method achieves scores of \textbf{153.4} for CIDEr and \textbf{78.2} for self-CIDEr, surpassing both BLIP and BLIP2. This shows the superiority of our concept-aware temporal modeling and set-prediction-based diversity modeling.

\subsection{Evaluation of Transferring Ability}
We evaluate the transferring ability of the diverse captioning models by running inference on MSVD with models trained on MSRVTT. As shown in Tab.\ref{tab:trans}, LSTM-based Div-BS and SCG-SP-LSTM perform poorly on relevance scores, indicating their inability to transfer the knowledge learned from one set to another. DML and SCG-SP-LSTM employ the powerful transformer-based captioner and have the transferring ability. Specifically, our SCG-SP-LSTM has better overall transferring performance than DML, verifying the benefit of using external knowledge.

\begin{table}
	\begin{center}	
		\resizebox{\columnwidth}{!}{
			\setlength{\tabcolsep}{2mm}{
				\small
				\begin{tabular}{lcccccccc}
					\toprule[1.5pt]
					\multirow{2}{*}{Models} & 
					\multicolumn{4}{c}{MSVD} & 
					\multicolumn{4}{c}{MSRVTT} \\
					\cline{2-9}
					& B@4 & M & R-L & C & B@4 & M & R-L & C  \\
					\midrule[0.5pt]
					Div-BS 
					& 14.5 & 31.6 & 57.9 & 55.4
					&  10.3 & 24.5 & 45.1 & 27.7 \\
					
					DML
					& 34.6 & 39.1 & 70.1 & 87.2
					& \textbf{17.2} & 26.1 & 53.2 & 31.2 \\
					\rowcolor{lightgray}
					SCG-SP-LSTM
					& \textbf{35.7} & \textbf{45.8} & \textbf{70.3} & \textbf{96.0}
					& 16.7 & \textbf{28.6} & \textbf{53.9} & \textbf{35.4} \\
					\rowcolor{lightgray}
					SCG-SP-Prefix
					& 33.6 & 43.9 & 69.5 & 88.6 
					& 16.4 & 27.9 & 52.4 & 34.5 \\
					\bottomrule[1.5pt]
				\end{tabular}
		}}
	\end{center}
	\caption{The average relevance scores of consensus top-5 captions on MSVD and MSRVTT.}
	\label{tab:consensus}
\end{table}

\begin{table}
	\begin{center}
		\resizebox{\columnwidth}{!}{
			\setlength{\tabcolsep}{0.8mm}{
				\small
				\begin{tabular}{lllcccccccc}
					\toprule[1.5pt]
					Model & Vis. Enc. & Lan. Dec.
					& B@4 & M & R-L & C  & D-1 & D-2 & m-B $\downarrow$ & s-C \\
					\midrule[0.5pt]
					BLIP & ViT-B & BLIP-text(0.16B)
					& 63.4 & 62.0 & 83.9 & 141.3 
					& 26.0 & 42.3 & 52.6 & 75.5 \\
					BLIP2 & BLIP2-vision & OPT(2.7B)
					& 57.2 & 60.4 & 81.7 & 141.7 
					& 23.3 & 38.7 & 74.9 & 75.8 \\
					\midrule[0.5pt]
					\multirow{3}{*}{Ours} & ResNet101 & GPT2(0.12B) 
					& 56.5 & 58.9 & 81.0 & 114.7
					& 24.3 & 43.1 & 58.4 & 67.9 \\
					& CLIP-ViT-L & GPT2(0.12B) 
					& 67.4 & 64.9 & 86.0 & 142.6
					& 29.9 & 51.8 & 39.8 & 74.2 \\
					& CLIP-ViT-L & LLaMA(1.3B)
					& \textbf{70.3} & \textbf{68.0} & \textbf{87.2} & \textbf{153.4} 
					& \textbf{30.5} & \textbf{51.3} & \textbf{39.2} & \textbf{78.2} \\
					\bottomrule[1.5pt]
				\end{tabular}
		}}
	\end{center}
	\caption{Scores of different models on MSVD.}
	\label{tab:llama}
\end{table}

\begin{table}[t!]
	\begin{center}
		\resizebox{\columnwidth}{!}{
			\setlength{\tabcolsep}{1.2mm}{
				\small
				\begin{tabular}{lcccccccc}
					\toprule[1.5pt]
					\multirow{2}{*}{Model} 
					& \multicolumn{4}{c}{Relevance Scores} 
					& \multicolumn{4}{c}{Diversity Scores} \\
					\cline{2-9}
					& B@4 & M & R-L & C  & D-1 & D-2 & m-B $\downarrow$ & s-C \\
					\midrule[0.5pt]
					Div-BS & 5.0 $\times 10^{-8}$ & 11.2 & 34.2 & 0.9 & 22.5 & \textbf{42.4} & \textbf{54.2} & \textbf{72.7} \\
					DML & \textbf{42.2} & 41.8 & 74.4 & 82.9 & 18.6 & 33.4 & 73.4 & 59.7 \\
					\rowcolor{lightgray}
					SCG-SP-LSTM & 3.3 $\times 10^{-8}$ & 11.6 & 34.7 & 1.0 & 16.7 & 27.9 & 77.2 & 52.8 \\
					\rowcolor{lightgray}
					SCG-SP-Prefix & 41.8 & \textbf{47.5} & \textbf{74.5} & \textbf{84.6} & \textbf{23.3} & \underline{39.3} & \underline{65.1} & \underline{63.3} \\
					\bottomrule[1.5pt]
				\end{tabular}
		}}
	\end{center}
	\caption{Relevance and diversity scores on MSVD with models trained on MSRVTT.}
	% \vspace{-3mm}
	\label{tab:trans}
\end{table}

\begin{figure}[t!]
	\begin{center}
		\includegraphics[width=\linewidth]{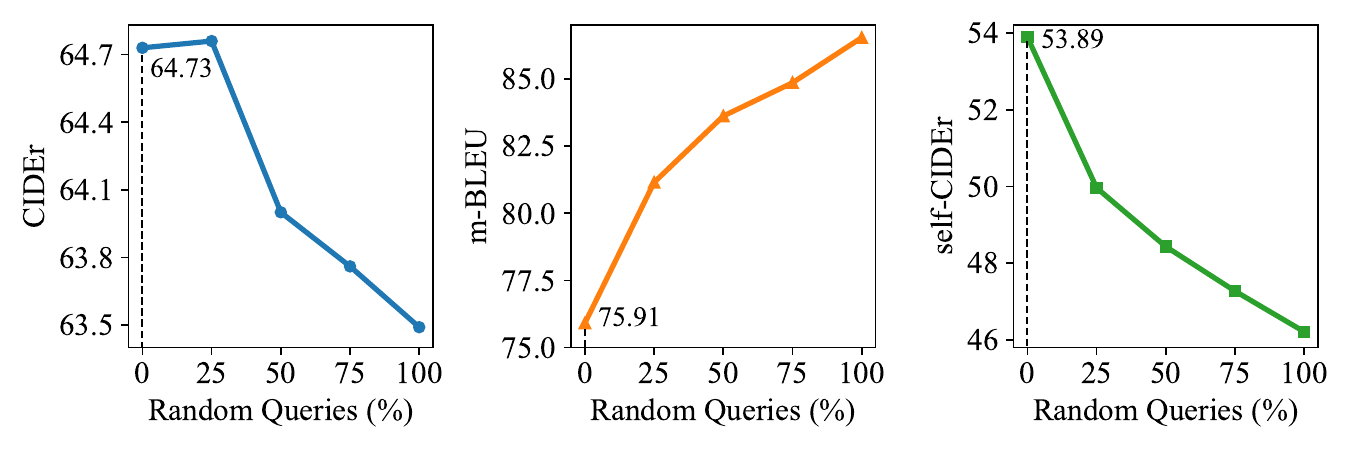}
	\end{center}
	\caption{Ablative results of the conceptual query. We replace part of conceptual queries with random vectors and plot score curves.
	}
	\label{fig:1_app}
\end{figure}

\subsection{Ablative Results}

\subsubsection{Conceptual Query}
In the main paper, we demonstrate that the semantic concept guidance benefits model performance. As an aspect of semantic concept guidance, the conceptual query encodes rich semantic information from visual content. We further evaluate the effectiveness of the conceptual query by replacing part of the input sequence of queries with random vectors. As shown in Fig.\ref{fig:1}, we plot the curves of CIDEr, m-BLEU ,and self-CIDEr when varying the percentage of the random queries. As the percentage of the random queries increases, both relevance and diversity performance drop. These results further prove the effectiveness of the conceptual query.

\subsubsection{Capacity of the Ground-Truth Set}
For each video, we build the ground-truth set with a capacity of $M'$ by clustering and sampling. We first cluster captions for each video into $M'$ samples by their sentence embeddings obtained from SBERT \cite{reimers2019sentence}. Then, at every training epoch, we randomly sample one caption from each of the $M'$ clusters to form the ground-truth set.

Tab.\ref{tab:1_app} shows the diversity statistics of the ground-truth set with different set capacities. The original $M'$ is 41 (on average) and 20 for MSVD and MSRVTT respectively without sampling. For comparison, we further set the $M'$ to half of the original values by sampling, i.e. 20/10 for MSVD/MSRVTT. For the sampled datasets, we report the average scores of 10 times of random sampling. Note that 

Compared with MSVD, MSRVTT owns a corpus with higher diversity. Sampling helps to increase the diversity of the ground-truth set. However, a lower set capacity does not guarantee the performance gain of the model. Tab.\ref{tab:2_app} shows that for MSVD, SCG-SP-LSTM performs better when trained on the sampled dataset with $M'$ set to 20. The case is the opposite on MSRVTT. We conclude that for a simple dataset like MSVD, sampling removes the redundancy of the dataset and helps with the model performance. For MSRVTT with distinctive ground-truth captions, sampling leads to the loss of valid information and causes a drop in relevance performance. Therefore, we only apply sampling on MSVD and fix the ground-truth set capacity to 20 for both datasets.

\begin{table}
	\begin{center}
		\resizebox{\columnwidth}{!}{
			\setlength{\tabcolsep}{2.5mm}{
				\begin{tabular}{llcccc}
					\toprule[1.5pt]
					Dataset & $M'$ & D-1 & D-2 & m-B $\downarrow$ & s-C \\
					\midrule[0.5pt]
					\multirow{2}{*}{MSVD} & $41^*$ & 21.8 & 45.9 & 46.4 & 70.2 \\
					& 20 (sampled) & 34.8 & 64.2 & 22.8 & 81.5 \\
					\midrule[0.5pt]
					\multirow{2}{*}{MSRVTT} & 20 & 37.5 & 69.0 & 25.3 & 84.0 \\
					& 10 (sampled) & 54.4 & 85.5 & 4.8 & 93.5 \\
					\bottomrule[1.5pt]
				\end{tabular}
		}}
	\end{center}
	\caption{The diversity statistics of ground-truth caption sets with respect to different datasets and capacities. $^*$ indicates the average capacity across all sets.}
	\label{tab:1_app}
\end{table}

\begin{table}
	\begin{center}
		\resizebox{\columnwidth}{!}{
			\setlength{\tabcolsep}{1.5mm}{
				\small
				\begin{tabular}{llcccccccc}
					\toprule[1.5pt]
					\multirow{2}{*}{Dataset} 
					& \multirow{2}{*}{$M'$} 
					& \multicolumn{4}{c}{Relevance Scores} 
					& \multicolumn{4}{c}{Diversity Scores} \\
					\cline{3-10}
					& & B@4 & M & R-L & C  & D-1 & D-2 & m-B $\downarrow$ & s-C \\
					\midrule[0.5pt]
					\multirow{2}{*}{MSVD} & $41^*$
					& 44.1 & 48.4 & 77.1 & 105.8
					& 9.0 & 13.5 & 91.3 & 29.1 \\
					& 20
					& \textbf{52.0} & \textbf{56.3} & \textbf{80.3} & \textbf{107.8}
					& \textbf{15.0} & \textbf{24.7} & \textbf{75.0} & \textbf{46.2} \\
					\midrule[0.5pt]
					\multirow{2}{*}{MSRVTT} & 20
					& \textbf{51.1} & \textbf{45.4} & \textbf{71.2} & \textbf{64.7}
					& 17.3 & \textbf{29.3} & \textbf{75.9} & \textbf{53.9} \\
					& 10
					& 49.7 & 44.3 & 70.5 & 63.4
					& \textbf{17.4} & \textbf{29.3} & 77.3 & 53.8 \\
					\bottomrule[1.5pt]
				\end{tabular}
		}}
	\end{center}
	\caption{Ablative results of the ground-truth set capacity. We train SCG-SP-LSTM on datasets of different ground-truth set capacities. Both relevance and diversity scores are reported. $^*$ indicates the average capacity across all sets.}
	% \vspace{-3mm}
	\label{tab:2_app}
\end{table}

\subsubsection{Prefix Length}
\begin{figure}[t!]
	\begin{center}
		\includegraphics[width=0.95\linewidth]{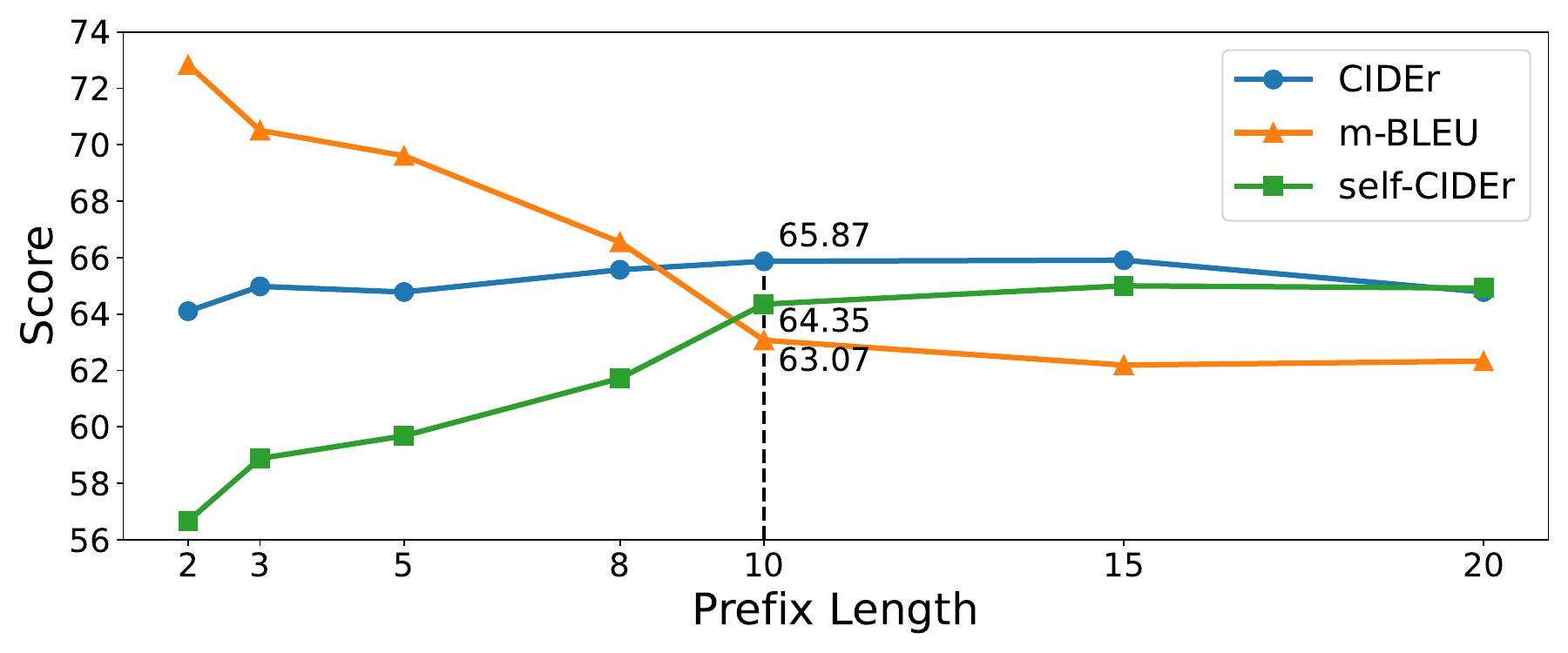}
	\end{center}
	\caption{Effect of prefix length.
	}
	\label{fig:plen}
\end{figure}

We set different prefix length of SCG-SP-Prefix and report scores including CIDEr, m-BLEU, and self-CIDEr. As shown in Fig.\ref{fig:plen}, as the prefix length increase, the performance of the model first improves and then saturates at 10. Further increasing prefix length does not bring obvious performance gain, but leads to higher computation load. Thus we set the prefix length to 10 for efficient training.

\subsubsection{Loss Weights}
Tab.\ref{tab:3_app} and Tab.\ref{tab:4_app} show the model performance under different settings of the weight of the classification loss $\lambda$ and the weight of the diversity regularization term $\lambda_d$. Note that $\lambda_d$ is set to 0.5 for experiments reported in Tab.\ref{tab:3_app} and $\lambda$ is set to 1 for experiments reported in Tab.\ref{tab:4_app}. We choose the setting with the best performance as $\lambda=1$ and $\lambda_d=0.5$.

\section{Appendix B: Additional Qualitative Results}
\begin{figure*}[p]
	\begin{center}
		\includegraphics[width=\linewidth]{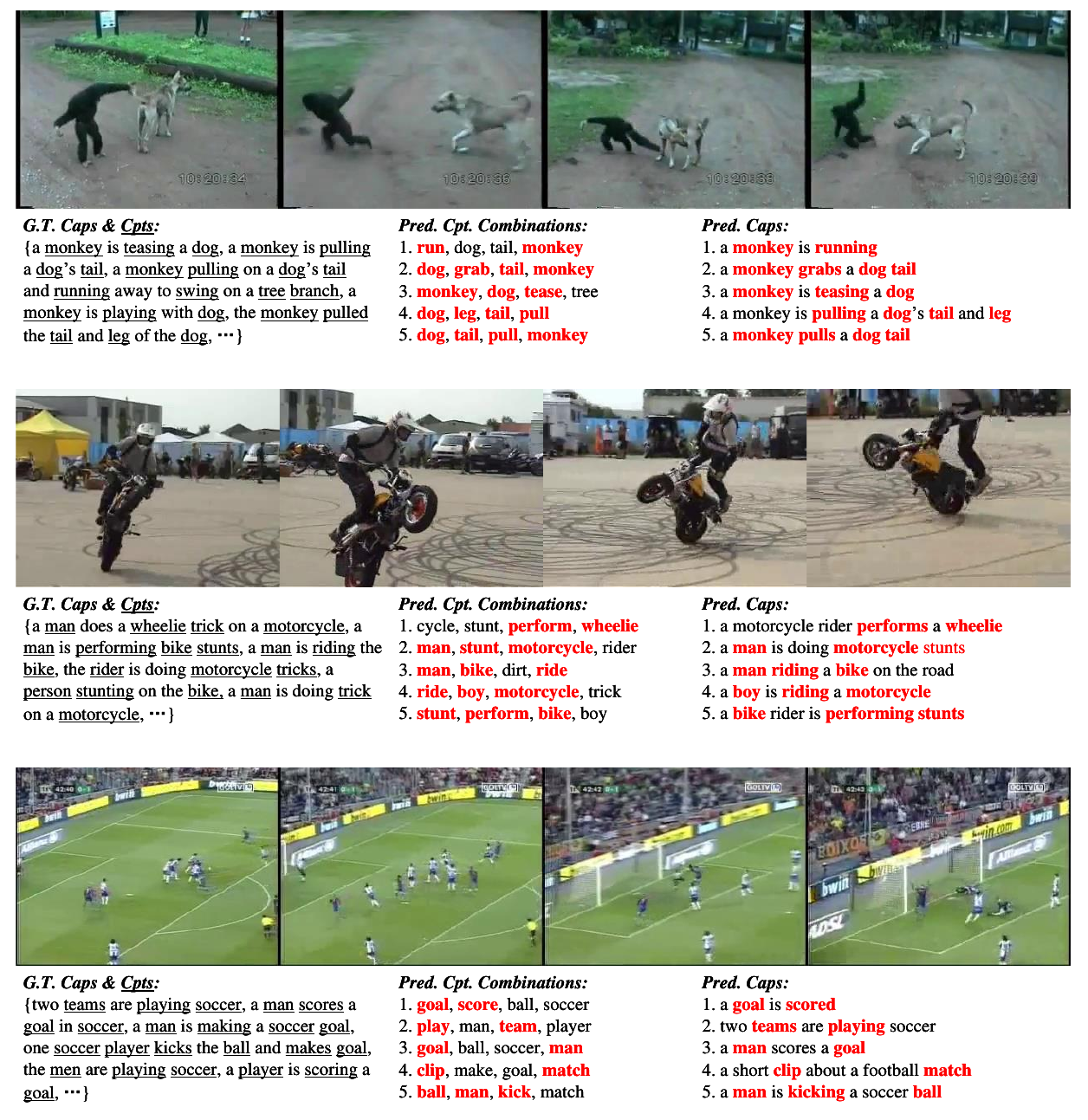}
	\end{center}
	\caption{An example of generations by SCG-SP-Prefix on MSVD. Concepts in ground-truth captions are underlined. Concepts both showed in predicted combinations and captions are highlighted in red. 
	}
	\label{fig:2_app}
\end{figure*}

\begin{figure*}[p]
	\begin{center}
		\includegraphics[width=\linewidth]{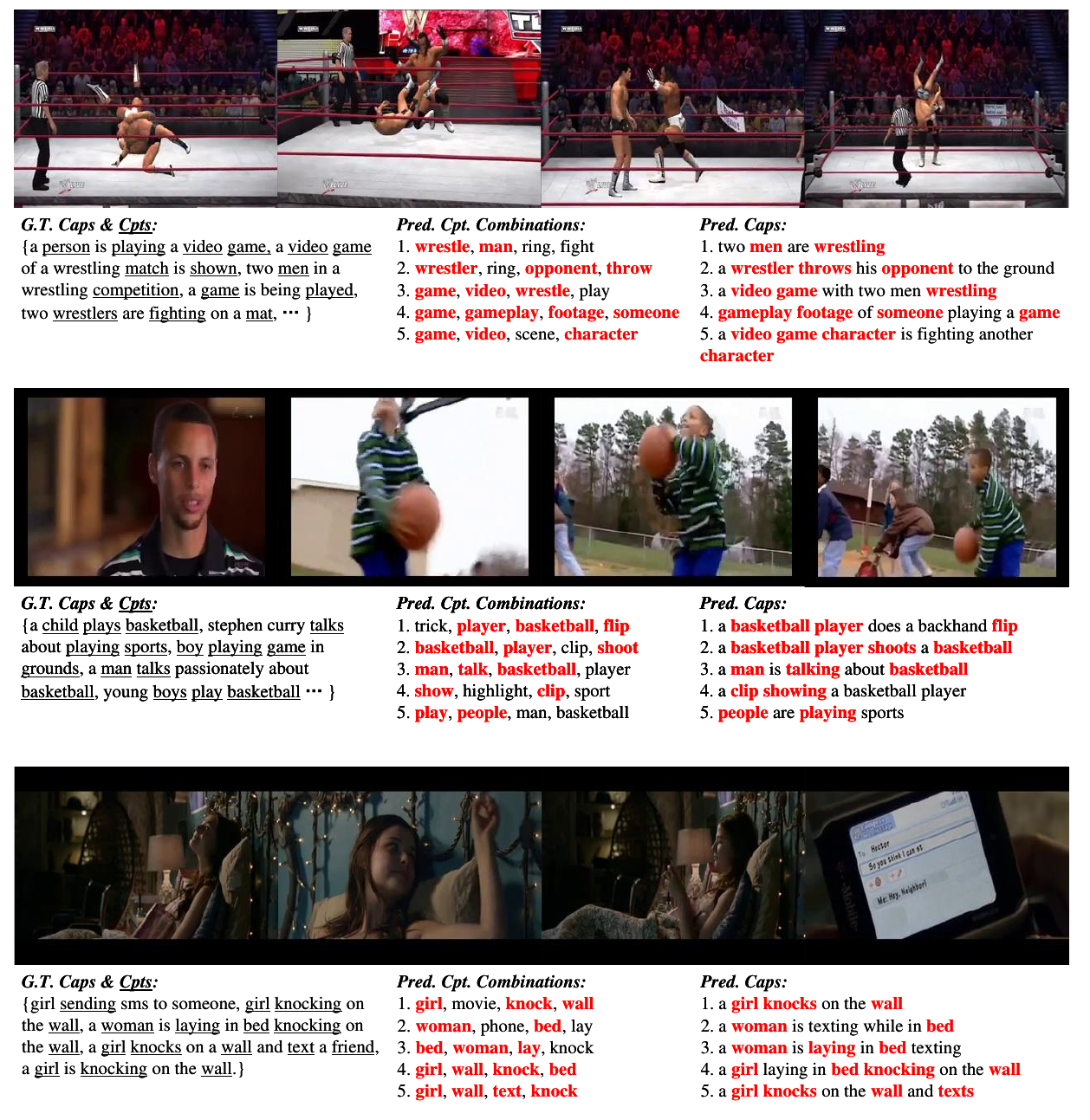}
	\end{center}
	\caption{An example of generations by SCG-SP-Prefix on MSRVTT. 
		Concepts in ground-truth captions are underlined. Concepts both showed in predicted combinations and captions are highlighted in red. 
	}
	\label{fig:3_app}
\end{figure*}

Fig.\ref{fig:2_app} and Fig.\ref{fig:3_app} show more examples of generations by our proposed SCG-SP-Prefix. Below the video frames, the ground-truth captions are listed with the concepts underlined. We present the predicted captions and corresponding concept combinations. Note that concepts both showed in predicted combinations and captions are highlighted in red.
The generated diverse captions are highly related to the predicted concept combinations. The results demonstrated the effectiveness and interpretability of our proposed model.

\begin{table}[t!]
	\begin{center}
		\resizebox{\columnwidth}{!}{
			\setlength{\tabcolsep}{1.5mm}{
				\small
				\begin{tabular}{ccccccccc}
					\toprule[1.5pt]
					\multirow{2}{*}{$\lambda$} 
					& \multicolumn{4}{c}{Relevance Scores} 
					& \multicolumn{4}{c}{Diversity Scores} \\
					\cline{2-9}
					& B@4 & M & R-L & C  & D-1 & D-2 & m-B $\downarrow$ & s-C \\
					\midrule[0.5pt]
					0.1 
					& 46.3 & 43.6 & 68.4 & 62.2
					& 11.8 & 17.8 & 91.0 & 36.8 \\
					0.5
					& 53.2 & 46.4 & 71.7 & 66.4
					& 18.9 & 32.1 & 72.1 & 57.8 \\
					\textbf{1}
					& \textbf{54.5} & \textbf{47.2} & \textbf{72.4} & \textbf{67.0}
					& \textbf{21.4} & \textbf{37.0} & \textbf{65.1} & \textbf{62.6} \\
					2
					& 54.2 & 46.6 & 72.2 & 66.1
					& 20.0 & 34.1 & 68.1 & 59.9 \\
					10
					& 51.9 & 45.7 & 71.5 & 64.1
					& 11.9 & 32.3 & 70.0 & 59.0 \\     
					\bottomrule[1.5pt]
				\end{tabular}
		}}
	\end{center}
	\caption{Ablative results of the weight of the classification loss $\lambda$.}
	% \vspace{-3mm}
	\label{tab:3_app}
\end{table}

\begin{table}[t!]
	\begin{center}
		\resizebox{\columnwidth}{!}{
			\setlength{\tabcolsep}{1.5mm}{
				\small
				\begin{tabular}{lcccccccc}
					\toprule[1.5pt]
					\multirow{2}{*}{$\lambda_d$} 
					& \multicolumn{4}{c}{Relevance Scores} 
					& \multicolumn{4}{c}{Diversity Scores} \\
					\cline{2-9}
					& B@4 & M & R-L & C  & D-1 & D-2 & m-B $\downarrow$ & s-C \\
					\midrule[0.5pt]
					0 (Base) 
					& 51.1 & 45.4 & 71.2 & 64.7
					& 17.3 & 29.3 & 75.9 & 53.9 \\
					0.05
					& 51.6 & 45.7 & 71.3 & 65.7
					& 17.9 & 30.5 & 74.4 & 55.8 \\
					\textbf{0.5}
					& \textbf{54.5} & \textbf{47.2} & \textbf{72.4} & \textbf{67.0}
					& \textbf{21.4} & \textbf{37.0} & \textbf{65.1} & \textbf{62.6} \\
					5
					& 44.4 & 42.7 & 68.7 & 59.4
					& 11.7 & 17.2 & 91.3 & 35.7 \\
					\bottomrule[1.5pt]
				\end{tabular}
		}}
	\end{center}
	\caption{Ablative results of the weight of the diversity regularization term $\lambda_d $.}
	% \vspace{-3mm}
	\label{tab:4_app}
\end{table}

\end{document}